\documentclass[runningheads]{llncs}
\usepackage{lineno}
\usepackage[T1]{fontenc}
\usepackage{graphicx}
\usepackage{algorithm}
\usepackage{algpseudocode}
\usepackage{amsmath}
\usepackage{amssymb}
\usepackage{orcidlink}
\usepackage[misc]{ifsym}
\usepackage{xcolor}
\usepackage{listings}

\begin{document}

\title{ExpliCIT-QA: Explainable Code-Based Image Table Question Answering}




\author{Maximiliano Hormazábal Lagos \inst{1,2}  \textsuperscript{(\Letter)} 
 \orcidlink{0009-0003-3687-1924} \and
Álvaro Bueno Sáez \inst{2} \orcidlink{0009-0006-9199-972X} \and
Pedro Alonso Doval \inst{2} \orcidlink{0009-0000-8255-3466} \and
Jorge Alcalde Vesteiro \inst{2} \orcidlink{0009-0003-6418-3694} \and
Héctor Cerezo-Costas \inst{2} \orcidlink{0000-0003-2813-2462}}

\authorrunning{M. Hormazábal et al.}

\institute{
Computer Vision Center, Universitat Autònoma de Barcelona, Spain \\
\email{mhormazabal@cvc.uab.es} \and
Gradiant, Vigo, Galicia, Spain \\
\email{\{abueno,hcerezo,palonso,jalcalde\}@gradiant.org}
}

\maketitle

\begin{abstract}
We present ExpliCIT-QA, a system that extends our previous MRT approach for tabular question answering into a multimodal pipeline capable of handling complex table images and providing explainable answers. ExpliCIT-QA follows a modular design, consisting of: (1) Multimodal Table Understanding, which uses a Chain-of-Thought approach to extract and transform content from table images; (2) Language-based Reasoning, where a step-by-step explanation in natural language is generated to solve the problem; (3) Automatic Code Generation, where Python/Pandas scripts are created based on the reasoning steps, with feedback for handling errors; (4) Code Execution to compute the final answer; and (5) Natural Language Explanation that describes how the answer was computed. The system is built for transparency and auditability: all intermediate outputs, parsed tables, reasoning steps, generated code, and final answers are available for inspection. This strategy works towards closing the explainability gap in end-to-end TableVQA systems. We evaluated ExpliCIT-QA on the TableVQA-Bench benchmark, comparing it with existing baselines. We demonstrated improvements in interpretability and transparency, which open the door for applications in sensitive domains like finance and healthcare where auditing results are critical. Code
available at \url{https://github.com/maxhormazabal/ExpliCIT}.

\end{abstract}
%
%
\section{Introduction}

Real-world documents often contain important information in tables, which are frequently embedded as images in PDF's. Tables may use complex layouts (with merged cells, multi-row headers, or irregular schemas). Question Answering over Tabular Data (TableQA) has seen progress when tables are provided in structured text in a stable row and column distribution. However, the complex visual tables on real documents usually have more uncommon structures with columns that overlap, multiple tables in one, etc. This scenarios still remains challenging in Visual Question Answering over Table Images (TableVQA). Vision-Language Models (VLLMs) like GPT-4V \cite{openai2024gpt4technicalreport} have demonstrated an impressive ability to directly interpret table images and answer questions, achieving the highest accuracy on benchmarks like TableVQA-Bench \cite{kim2024tablevqabenchvisualquestionanswering}. However, these end-to-end models act as black boxes: they do not reliably reveal how the answer was obtained or which table cells were used, and their internal reasoning is hidden and often ambiguous. In high-stakes domains like finance or healthcare, such lack of transparency is problematic, as users need to trust and verify the answer derivation \cite{lakkaraju2023llms,nazi2024large}.

That is why explainability in current TableVQA systems is an open gap to explore in both classical pipelines and end-to-end VLLMs to offer different levels of traceability in their decision process. This can be done by explaining predictions from end-to-end models by the understanding of their attention maps \cite{bracsoveanu2022visualizing,park2018multimodal} and also by delegating the final decision to a simpler model or procedure that can be easily explainable or even interpretable. Recent work has begun addressing this gap with proposals like Plan-of-SQLs (POS) \cite{nguyen2025interpretablellmbasedtablequestion}, which decomposes table queries into interpretable sub-queries (SQL statements) to explain the reasoning. Their approach allowed verification of answers, without sacrificing accuracy on text-table QA. TRH2TQA for table images performs first detailed table recognition and then feeds the structured data to a QA model \cite{10944147}.

We propose ExpliCIT-QA, an extension of our Maximizing Recovery from Tables (MRT) \cite{hormazabal-et-al} to deal with complex table images in a multi-step VLLMs and Large Language Models (LLMs) to offer a natural language explanation based of how the answer was calculated. The key idea is to convert and transfer this tabular data pipeline to the task of visual question answering on complex table images by a series of transparent sub-tasks: (1) re-interpret a table image with a complex layout to transform it into serialized table data by Chain-of-Thought (CoT) reasoning (2) obtain and structure a step-by-step explanation of how to calculate the answer to the question, (3) translate the reasoning into executable python code, (4) compute the answer with that code and (5) get a final explanation and description of the actual step carried out to get the answer which can be examined by users.

This work makes the following contributions:
\begin{itemize}
    \item \textbf{ExpliCIT-QA}: A novel pipeline that extends MRT approach for visual question answering for table images by integrating VLM and LLM reasoning.
    \item \textbf{Explainability and Traceability}: We highlight the importance of transparency in TableVQA. Our system provides human-readable CoT explanations and runnable code for every answer, addressing the ambiguity of end-to-end model explanations
    \item \textbf{Evaluation on TableVQA-Bench}: We present experiments on the TableVQA-Bench dataset, comparing ExpliCIT-QA with baselines.
\end{itemize}

This paper is organized as follows. Section 2 reviews related work in TableQA, TableVQA, CoT, and program-aided reasoning. Section 3 details the ExpliCIT-QA system design. Section 4 presents experimental results and comparisons. Section 5 offers discussions on accuracy vs. interpretability and future improvements, and Section 6 concludes the paper.

\section{Related Work}


\subsection{Question Answering over Tabular Data}
Traditional Table QA assumes a structured table as input in order to read table content directly. Recent large-scale datasets such as WikiTableQuestions \cite{DBLP:journals/corr/PasupatL15}, WikiSQL \cite{DBLP:journals/corr/abs-1709-00103} and TabFact \cite{DBLP:journals/corr/abs-1909-02164} have driven progress in textual table QA. In particular, LLMs fine-tuned for table tasks \cite{DBLP:journals/corr/abs-2004-02349,DBLP:journals/corr/abs-2005-08314} and prompt-based methods \cite{lan2023improving} have achieved high accuracy. However, as we mentioned above, these solutions often struggle with traceability and explanations.

\subsection{Table Visual Question Answering}

Visual Table QA (TableVQA) extends this challenge where the table is given as an image, requiring visual understanding. Early attempts decomposed this into pipeline stages: detect table region, perform OCR, and then apply a textual QA model. Such pipelines were interpretable but prone to OCR errors and required significant engineering for each format. End-to-end deep models \cite{hu2024mplugdocowl2highresolutioncompressingocrfree,DBLP:journals/corr/abs-2111-15664,DBLP:journals/corr/PasupatL15,DBLP:journals/corr/abs-1909-02164,DBLP:journals/corr/abs-1911-10683} have been proposed to directly answer from document images without explicit structure extraction but internal workings remain opaque. 

VLLMs struggle with complex tables unless multiple vision queries are allowed. For instance, TableVQA-Bench \cite{kim2024tablevqabenchvisualquestionanswering} results showed that GPT-4V had 54.5\% accuracy with a vision-language only approach whereas its performance increase to 60.7\% when separating the inference process in two steps with multi-modal Table Structure Reconstruction followed by a LLM based inference. But it still underperformed when compared to an LLM that was given an equivalent text-input of the table itself. 

While these strategies have shown progress in the task of TableVQA, they mostly rely on black box models that do not favor explainability and provide an opportunity to improve reliability by explicitly extracting table structure first, and then reasoning over it. Pipeline approaches for document QA that combine multiple deep learning modules are more transparent and monitored \cite{nguyen2025interpretablellmbasedtablequestion,10944147,wang2024chainoftableevolvingtablesreasoning,wu2025protrixbuildingmodelsplanning}.

\subsection{Chain-of-Thought Reasoning}

CoT encourages LLMs to "think aloud" and produce intermediate reasoning steps \cite{wei2022chain}. It has been shown to improve performance on arithmetic, logical, and multi-hop reasoning tasks, by breaking problems into manageable steps \cite{mitra2024compositional,li2025structured}. Current state-of-the-art models can operate in a thorough reasoning mode, performing self-consistency checks and detailed step-by-step derivations, at the cost of some latency \cite{guo2025deepseek,yang2025qwen3}. Strategies such as \cite{CoCoNut} suggest that thinking can be done in the form of patches or latent knowlege, however, the standard is still to think in the form of text. This means that CoT can also be partially applied to VLLMs in the textual step of the generation that is to be used for interpretation of visual elements and their spatial distribution.

\subsection{Code-based Answering}

A way to ensure faithful reasoning is to delegate problem solving (in this case answering the question) to a simpler and interpretable model or algorithm based on the LLM output. Most of the table data solutions opt to generate executable code as the intermediate reasoning step showing that this process can yield more reliable and explainable results. This approach follows the Program-aided Language Models (PAL) \cite{gao2023palprogramaidedlanguagemodels} paradigm where the LLM's job is to produce a correct program that a computer can execute to get the answer.

The code serves as a formal specification of the reasoning eliminating ambiguity and arithmetic mistakes by delegating those to a Python/SQL interpreter system. The results of the system can be easily monitored, being able to generate explanations of how the output was computed and providing traceability as well. Furthermore, they do not need to process all the table structure at once, as some of the end-to-end solutions do, thus sidestepping problems related to context window size with large tables.

\section{Methodology}

\begin{figure}[h!]
\centering
\includegraphics[width=\textwidth]{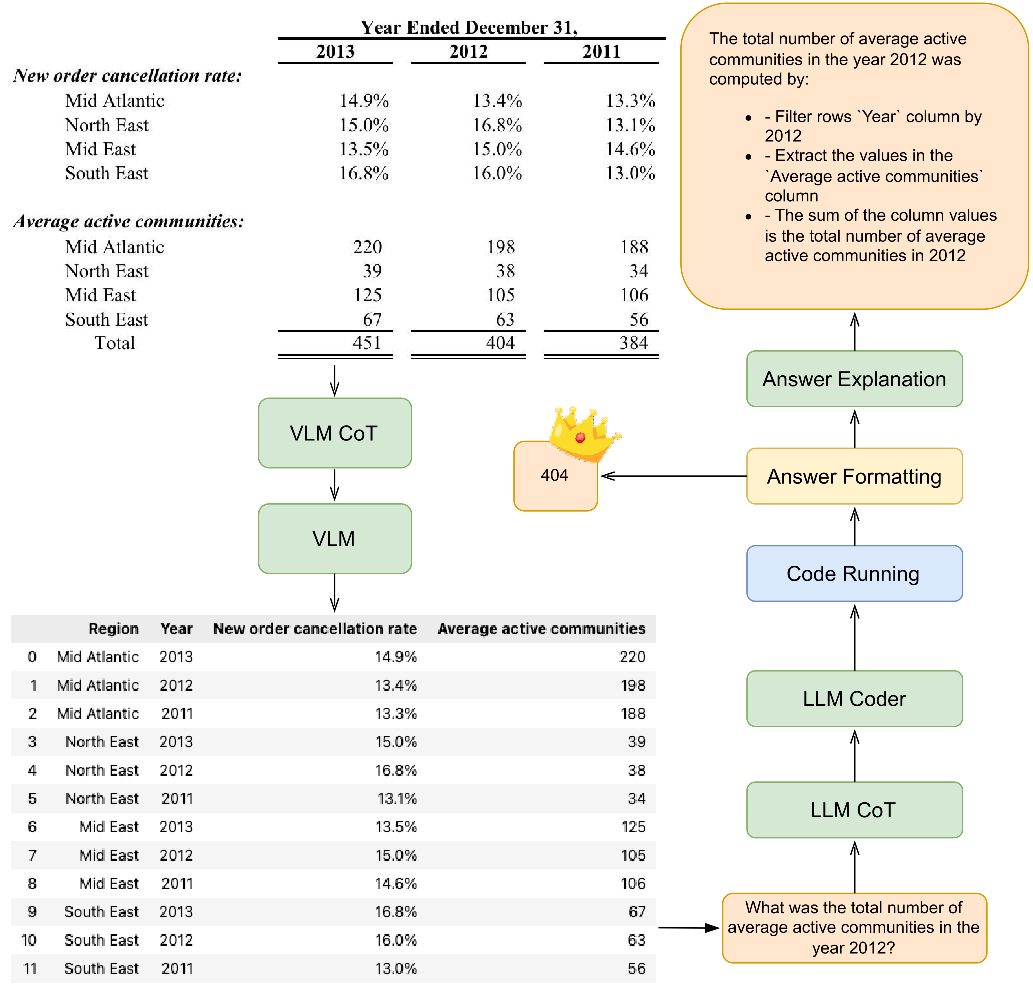}
\caption{ExpliCIT-QA: (1) Multimodal understanding: a vision model generates a visual reasoning to extract and structure the tabular content of a complex layout table image; (2) Natural language reasoning: an LLM produces a human-readable step-by-step describing the logic of the answer; (3) Automatic code generation: the steps are translated into executable code; (4) Code execution: the script is run deterministically on the extracted table to compute the answer; and (5) Final answer and explanation: how each operation in the code resulted in the result is synthesized in natural language.} \label{fig:pipeline}
\end{figure}

Our system consists of five main modules that operate sequentially: (1) Multimodal Table Understanding, (2) Language-based Reasoning based on CoT, (3) Automatic Code Generation with error feedback, (4) Code Execution and (5) Natural Language Explanation. Figure \ref{fig:pipeline} conceptually illustrates the pipeline. All intermediate steps such as extracted table, model propmts, reasoning, code and explanation are available as auditable outputs, allowing to manually verify the traceability of the process.

Given an input consisting of a table image and a natural language question, the following steps are
performed:

\subsection{Multi-modal Table Understanding}

The table image is processed to extract its content along with the table structure. This extraction is not a simple parsing process because the target data are tables with a complex layout usually composed of multiple sub-tables joined together for enterprise reports. A simple parsing process is not always going to be able to be easily translated into a tabular data format like \textit{csv}.  We employ Qwen-2.5-VL \cite{bai2025qwen2} for this task because of its powerful document understanding capabilities.

We decompose this step into two substeps. First we use the VLLM spatial reasoning to generate a CoT of the 'to-do inner steps' and considerations to transform this complex layout into a stable two-dimensional tabular data that can be parsed into a formal structure. Secondly, we use the 'to-do list' generated on step one to use the the model to generate a \textit{csv} like data with the information of the table. This often involves replicating and redundant data in order to reduce the dimensionality of the table and make it more easily manageable.

This structured table can be directly verified by humans (or in the worst case scenario another VLLM acting as a judge) against the image, ensuring no critical data was lost or misread.

\subsection{Natural language instructions with Chain-of-Thought}

Following down the pipeline the question and the extracted table data are given to the reasoning module. We use the recently released Qwen 3 \cite{yang2025qwen3technicalreport} LLM known for its reasoning and multi-step problem solving abilities. We produce a step-by-step explanation of how to get the answer to a question using the table. The prompt format includes a sample of the table, the column name, the question, and instructions. The output is not only the steps in natural language but also important insights such as columns to be used and values to be filtered. We implemented helper functions that hat employ fuzzy search in case categorical values are slightly mismatched during any part of the pipeline. Real names are substituted for those wrongly generated by the LLM.

This CoT steps are human-readable and each step references specific table content or operations. Having the model articulate the logic helps with the validation that the model has correctly understood the question and table structure.

\subsection{Automatic Code Generation}

Once we have the reasoning steps, the next module converts these steps into an executable code using Python and Pandas as a dataframe library. We found that Qwen 3 is capable of generating correct Pandas code for most cases, especially in cases like this where tables were transformed to two-dimension format.

We emphasize that our code is not hand-written: the system generates it on the fly, which is feasible thanks to the reasoning trace as a guide. It also means the final answer is not just whatever the LLM guesses, but is the output of a deterministic computation, eliminating occasional arithmetic or logical mistakes made by LLMs

Similar to the fuzzy search helper functions, we have an stock of pre-made helper functions for data management that can be referenced by the model in this code generation process and have more accurate results.

\subsection{Code Execution and Error Feedback}

Generated code is executed in a Python environment with the DataFrame from step (1) available. It has a $max\textunderscore tries$ parameter set to $3$ in order to give the coder module the opportunity to solve possible errors by checking the error and re-generating the code with feedback. 

As we mentioned, in the code-generation step the executor has access to the helper functions and can re-use them to avoid common mistakes when they have to build them from scratch. Those function definitions are included in the prompt in this step along the question and natural language instructions. Once the code runs without exceptions, we obtain the final answer to the question.

\subsection{Explanation step}

Finally, once the answer is obtained, a natural language explanation is elaborated based entirely on the code that has been executed to compute the answer. To some extent the same explanation could be reused as a step-by-step description of how to get the answer, however, it is not strictly necessary that the code is always fully linked to the explanation, so we have decided to use a previously trained model to extract the code steps used in a python script \footnote{ \href{https://huggingface.co/sagard21/python-code-explainer}{huggingface.co/sagard21/python-code-explainer}} this brings us closer to the explanation being linked to the arithmetic and logical processes used to compute and obtain the answer.


The step-by-step behavior of ExpliCIT-QA, we present a complete example taken from the \textit{FinTabNetQA} subset of the TableVQA-Bench benchmark.

\textbf{Question:} \textit{What was the net sales for North America in the year 2013?}

\subsubsection{Visual Understanding} 

\begin{figure}[h!]
\centering
\includegraphics[width=1\textwidth]{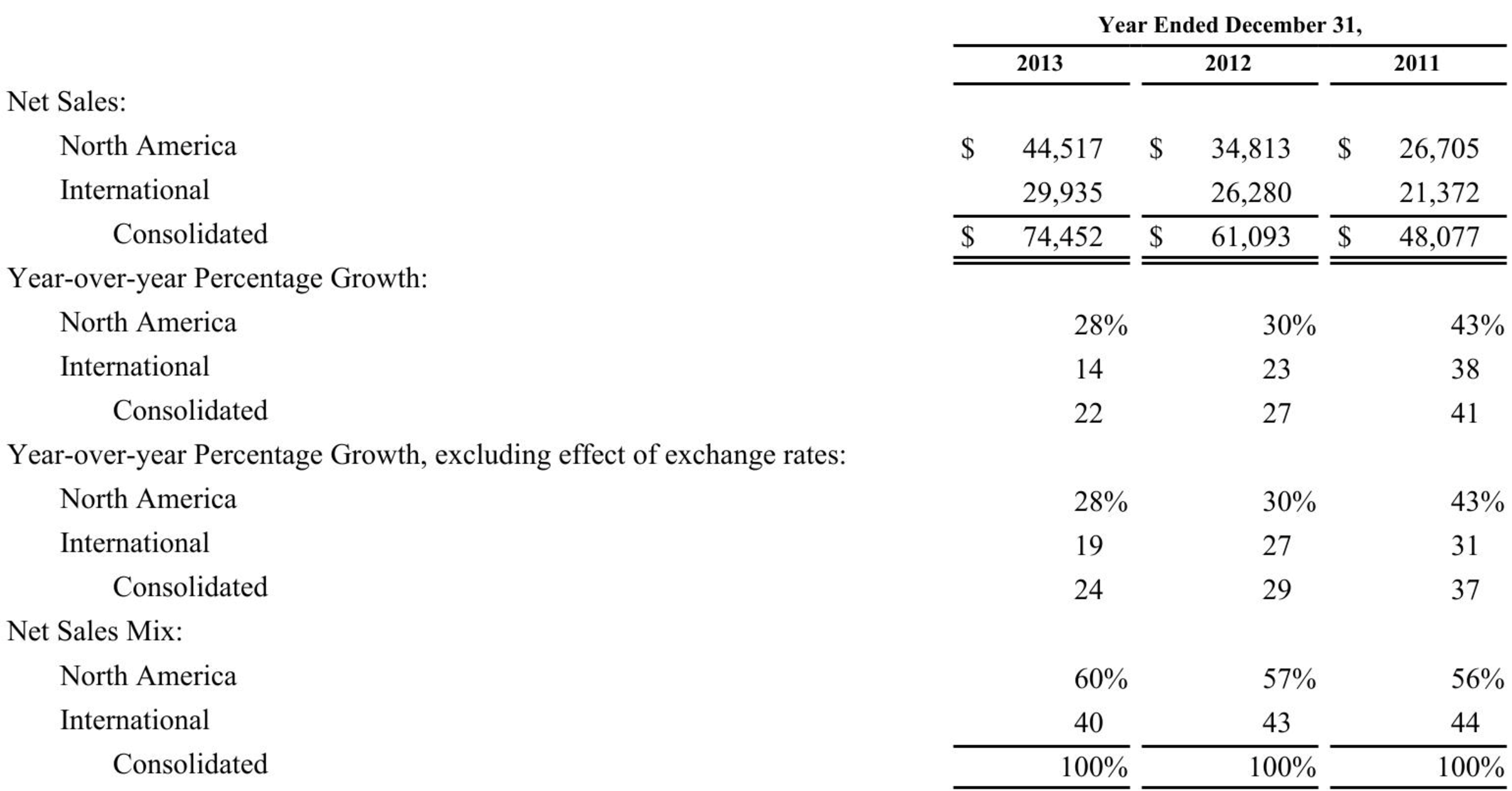}
\caption{Example of image which belongs to the FinTabNetQA subdataset on the TableVQA-Bench.} \label{fig:example}
\end{figure}

Following the example the table \ref{tab:example} shows the actual tabular data extracted for the image \ref{fig:example} which contains a complex financial layout with merged cells and multi-row headers. The system produces a visual reasoning trace that decomposes the layout into a normalized CSV format with the following columns: \texttt{Year}, \texttt{Region}, \texttt{Net Sales}, \texttt{YoY \% Growth}, \texttt{YoY \% Growth (ex FX)}, and \texttt{Net Sales Mix}.

\begin{table}[h!]
\centering
\caption{Output example of the resulting csv like tabular data after Table Understanding process.}
\begin{tabular}{|c|p{3cm}|p{2cm}|p{2cm}|p{2cm}|p{2cm}|}
\hline
\textbf{Year} & \textbf{Region} & \textbf{Net Sales} & \textbf{YoY \% Growth} & \textbf{YoY \% Growth (ex FX)} & \textbf{Net Sales Mix} \\
\hline
2011 & North America & \$26,705 & 43\% & 43\% & 56\% \\
\hline
2011 & International & \$21,372 & 38\% & 31\% & 44\% \\
\hline
2012 & North America & \$34,813 & 30\% & 30\% & 57\% \\
\hline
2012 & International & \$26,280 & 23\% & 27\% & 43\% \\
\hline
2013 & North America & \$44,517 & 28\% & 28\% & 60\% \\
\hline
2013 & International & \$29,935 & 14\% & 19\% & 40\% \\
\hline
\end{tabular}
\label{tab:example}
\end{table}

\subsubsection{Natural Language Reasoning}

The extracted table and the question are processed by Qwen 3, which generates the following CoT:

\begin{enumerate}
    \item Filter the table to include only rows where the \texttt{Region} column is equal to ``North America'' and the \texttt{Year} column is equal to 2013.
    \item Verify that the \texttt{Net Sales} column contains a numeric value for the filtered row.
    \item Retrieve the value from the \texttt{Net Sales} column for the filtered row.
    \item Ensure that no additional calculations or transformations are applied to the \texttt{Net Sales} value.
    \item Return the retrieved \texttt{Net Sales} value as the final answer.
\end{enumerate}

\subsubsection{Automatic Code Generation, Execution and Feedback} The reasoning is automatically translated into executable Pandas code that is executed within a controlled Python environment with error recovery mechanisms (up to 3 retries).

\begin{verbatim}
import pandas as pd

def parse_dataframe(df: pd.DataFrame) -> str:
    # <<< MAIN LOGIC IMPLEMENTATION START >>>
    # Step 1: Filter the DataFrame for 'North America' in 2013
    filtered_df = df[
    (df['Region'] == 'North America') & (df['Year'] == 2013)
    ]
    
    # Step 2: Verify that the 'Net Sales'
    # column contains a numeric value
    if not filtered_df.empty 
    and pd.api.types.is_numeric_dtype(filtered_df['Net Sales']):
        # Step 3: Retrieve the 'Net Sales' value
        net_sales_value = filtered_df['Net Sales'].values[0]
        # Step 4: Assign to result and cast to string
        result = str(net_sales_value)
    else:
        result = ""
    # <<< MAIN LOGIC IMPLEMENTATION END >>>
    
    return result
\end{verbatim}

\subsubsection{Final Explanation}

A natural language explanation is generated based on the actual procedure to compute the answer:

\begin{center}
    \textit{The net sales for North America in the year 2013 were calculated by filtering the table for that year and region, verifying the numeric value in the ‘Net Sales’ column, and extracting the result.}
\end{center}
\subsubsection{Result}
\begin{itemize}
    \item Ground truth: \texttt{\$44,517}
    \item Model output: \texttt{44517}
\end{itemize}

\noindent Although numerically correct, the output differs in format, which highlights the importance of normalization strategies for exact-match evaluation.

\section{Experimentation}

Our evaluation is conducted on the TableVQA-Bench dataset \cite{kim2024tablevqabenchvisualquestionanswering}, which consolidates several sub-datasets (VWTQ, VWTQ-Syn, VTabFact, and FinTabNetQA) targeting different table-centric reasoning capabilities in visual contexts. Each instance in the benchmark consists of a table image and a natural language question. Each domain represents different structural and contextual challenges in table images, ranging from synthetic layouts to financial reports. 

The pipeline evaluation consisted of two main stages: Table Understanding (TU) and Question Answering, which uses language-based reasoning and code execution to compute the final answer. For the visual table understanding, both the layout-aware CoT and cell extraction were performed using Qwen 2.5-VL 8B, a vision-language model known for robust multimodal understanding. For the language reasoning stage we have tested three configurations of the ExpliCIT-QA reasoning engine, using Qwen 3 at different parameter scales (4B, 8B, and 14B). Each version was responsible for generating the reasoning trace and executable Python code required to compute the answer based on the extracted table data.

Table \ref{tab:qa_accuracy} presents the results of our pipeline on the four sub-datasets, comparing three Qwen 3 configurations. For context, we include scores from the benchmark TableVQA-Bench paper where models use a similar two-stage pipeline to parse visual tables and then apply an LLM for question answering.

\begin{table}[h]
\centering
\caption{Accuracy comparison on TableVQA-Bench sub-datasets. Our results (Qwen-based pipeline) are compared to the state-of-the-art two-stage pipeline (GPT-4V $\rightarrow$ GPT-4).}

\begin{tabular}{|p{3cm}|c|c|c|c|c|c|}
\hline
\textbf{Model} & \textbf{VWTQ} & \textbf{VWTQ-Syn} & \textbf{VTabFact} & \textbf{FinTabNetQA} & \textbf{Avg.} \\ \hline
$\text{GPT-4V} + \text{GPT-4}$\cite{kim2024tablevqabenchvisualquestionanswering} & \textbf{45.2} & \textbf{55.6} & \textbf{78.0} & \textbf{95.2} & \textbf{60.7} \\ \hline
$\text{Gemini-ProV} + \text{Gemini-Pro}$\cite{kim2024tablevqabenchvisualquestionanswering} & 34.8 & 40.4 & 71.0 & 75.6  & 48.6 \\ \hline
T.U pipeline + Qwen3-4B & 35.99 & 44.97 & 2.68 & 29.76 & 27.69 \\ \hline
T.U pipeline + Qwen3-8B & 38.85 & 53.02 & 2.23 & 31.22 & 30.27 \\ \hline
T.U pipeline + Qwen3-14B & 42.99 & 49.66 & 4.91 & 29.76 & 31.50 \\ \hline
\end{tabular}
\label{tab:qa_accuracy}
\end{table}

The accuracy values across different Qwen 3 configurations remain comparable, showing only marginal variation. While our system does not yet reach the performance of top-tier vision-to-text pipelines like GPT-4V/GPT-4 or GeminiProV/GeminiPro, it maintains a competitive accuracy level given its primary design goal: maximizing explainability. The models used in our experimentation are much smaller than current state of the art models. Our strategy has very promising results and is model agnostic. Hence different models could be tested depending on the hardware limitations. Notably, our structured reasoning and executable code path enable interpretability on answer computation, a feature lacking in most end-to-end models.

We observe significantly higher accuracy on VWTQ and VWTQ-Syn across all Qwen configurations, which aligns with their simpler, Wikipedia-style structure and well-distributed QA design. For FinTabNetQA this subset contains financial tables with multi-row headers and merged cells. Despite using CoT for layout reconstruction, these structures remain challenging and prone to misinterpretation. In contrast, VTabFact suggests that fact verification tasks, especially those requiring semantic alignment rather than numeric computation, are less compatible with our current code-based reasoning approach.

Our current pipeline prioritizes transparency and step-wise auditability over raw performance. As a result, while the accuracy lags behind state-of-the-art LLM-backed vision QA systems, it provides a unique benefit in terms of trustworthiness, making it a promising direction for applications in regulated or high-stakes domains. We discuss these trade-offs and future optimization strategies for this system.

\section{Discussion}

Experiments on TableVQA-Bench reveal both positive aspects and areas for improvement that are worth examining closely. First, the performance gap between tables extracted directly from Wikipedia (VWTQ) and those generated synthetically (VWTQ-Syn) suggests that the models still rely on source-specific clues rather than on a deep understanding of the tabular structure. While for 14B configurations average accuracies close to 49\% are achieved in VWTQ-Syn, in FinTabNetQA these fall in several cases below 30\%. This dependence on graphical elements (colors, margins, fonts) shows that generalization outside the original domain is still limited.

The drastic drop in accuracy in VTabFact, where it does not exceed 4,9\%, reinforces the hypothesis that the main point of improvement of this system is the Table Understanding phase and that currently mechanisms to deal with irregular layouts such as multilevel headers, merged cells and footnotes cause recurrent errors in cell identification and cross-information aggregation.

Another critical point is the overall accuracy metric beyond this benchmarking. In many QA systems the accuracy metric is applied in an exact-match logic. While this is a clear and simple way to evaluate, it does not differentiate between formatting errors. It is true that this benchmarking proposes to use looser versions of accuracy (called \textit{relieved accuracy}) that use parsing and normalization processes to decide if the answers are correct or not, however, these are introduced in a customized way to each split of the dataset and this can lead to fundamental interpretation errors. An open gap is to extend the use of more general metrics such as Average Normalized Levenshtein Similarity (ANLS) proposed in the ICDAR 2019 \cite{DBLP:journals/corr/abs-1907-00490} Competition which also offers opportunities for improvement but can mean a more general comparison and not so directed to the specific dataset to evaluate QA systems.

\section{Conclusions and Future Work}
ExpliCIT-QA demonstrates that the combination of multimodal vision, reasoning, and code execution is viable for Visual Table QA, offering traceability and arithmetic error reduction. The accuracy gap in complex tables is a major constraint to extending its application in real environments with diverse formats, so improving the understanding of the more complex structures in table images is vital to improving this approach.

Future work will include different approaches to understand and simplify information from tables with distributions that include various dimensions of analysis, merged cells, complex layouts and others. This may involve sub-processing and normalization and denormalization in order to arrive at a simpler representation using a code-based approach that is in itself much more explainable and verifiable than end-to-end black-box methods. It is also a priority to be able to measure the internal coherence of the CoT and the fidelity of the generated explanations so that the final explanation makes even more sense and to dispense with extra code explanation processes by having a verifiable correlation between the steps to be followed and the executed code.

Other interesting avenues to explore are intermediate representations such as graphs to facilitate image conversion to coherent data structures. Evaluate hybrid methods combining neural networks with symbolic rules to better handle the variety of layouts.

The real lever for improvement will lie in strengthening the multimodal understanding of heterogeneous tabular structures and in diversifying the evaluation criteria to reflect not only accuracy, but also robustness and explanatory coherence. These efforts will help reduce latency, increase flexibility in the face of new table formats and layouts and close the gap with purely performance-oriented systems.

\subsubsection*{Disclosure of Interests}

This work was carried out by employees of Gradiant (Vigo, Spain) and a student from the Universitat Autònoma de Barcelona (UAB) as part of the UAB PhD program. The participation of the UAB student was supported by the National Agency for Research and Development (ANID) / Scholarship Program / BECAS CHILE / 2024-72240209, Government of Chile.

%

%
%
%
\bibliographystyle{splncs04}
\bibliography{Ref}

\end{document}